\documentclass{article}
\usepackage{ijcai17}

\usepackage{times}
\usepackage{times}
\usepackage{latexsym}
\usepackage{xspace}
\usepackage{color}
\usepackage{url}
\usepackage{graphicx}
\usepackage{multirow}
\usepackage{makecell}
\usepackage{amsmath}
\usepackage{amssymb}
\usepackage{subcaption}
\usepackage{bm}
\usepackage{algorithm} 
\usepackage[noend]{algpseudocode}
\usepackage{mathtools}
\mathtoolsset{showonlyrefs=true}
\usepackage{enumitem}
\title{End-to-end optimization of goal-driven and visually grounded dialogue systems}

\author{Florian Strub \\ \small florian.strub@inria.fr \\ \small Univ. Lille, CNRS, Centrale Lille, Inria, \\ \small  UMR 9189 - CRIStAL, F-59000 Lille, France\normalsize 
\And Harm de Vries \\ \small mail@harmdevries.com\normalsize \\ \small University of Montreal
\And Jeremie Mary \\ \small jeremie.mary@univ-lille3.fr\normalsize \\ \small Univ. Lille, CNRS, Centrale Lille, Inria, \\ \small  UMR 9189 - CRIStAL, F-59000 Lille, France\normalsize 
\AND 
Bilal Piot \\ \small piot@google.com\normalsize \\ \small DeepMind\normalsize  
\And Aaron Courville \\ \small aaron.courville@gmail.com\normalsize \\ \small University of Montreal
\And Olivier Pietquin \\ \small pietquin@google.com\normalsize \\ \small DeepMind\normalsize }
\makeatletter
\renewcommand{\paragraph}{%
  \@startsection{paragraph}{4}%
  {\z@}{1ex \@plus 1ex \@minus .2ex}{-1em}%
  {\normalfont\normalsize\bfseries}%
}
\makeatother

\newcommand{\Image}{\mathcal{I}}
\newcommand{\GW}{GuessWhat?!\xspace}
\newcommand{\yestoken}{{{<}yes{>}}}
\newcommand{\notoken}{{{<}no{>}}}
\newcommand{\natoken}{{{<}na{>}}}

\newcommand{\stoptoken}{{{<}?{>}}}
\newcommand{\stopdialoguetoken}{{{<}stop{>}}}

\newcommand{\ft}[2]{#1:#2}

\begin{document}

\maketitle

\begin{abstract}
End-to-end design of dialogue systems has recently become a popular research topic thanks to powerful tools such as encoder-decoder architectures for sequence-to-sequence learning. Yet, most current approaches cast human-machine dialogue management as a supervised learning problem, aiming at predicting the next utterance of a participant given the full history of the dialogue. This vision is too simplistic to render the intrinsic planning problem inherent to dialogue as well as its grounded nature, making the context of a dialogue larger than the sole history. This is why only chit-chat and question answering tasks have been addressed so far using end-to-end architectures. In this paper, we introduce a Deep Reinforcement Learning method to optimize visually grounded task-oriented dialogues, based on the policy gradient algorithm. This approach is tested on a dataset of 120k dialogues collected through Mechanical Turk and provides encouraging results at solving both the problem of generating natural dialogues and the task of discovering a specific object in a complex picture. 
% * <aaron.courville@gmail.com> 2017-02-20T01:25:00.788Z:
% 
% > making the context of a dialogue larger than the sole history.
% Also this? Sole history?
% 
% ^ <florian.strub@gmail.com> 2017-02-20T02:36:48.056Z:
% 
% Idem, I would suggest:
% Modeling the distribution only over the user utterances may miss the inherent contextual aspect of real life dialogues. 
% 
% ^.
% * <aaron.courville@gmail.com> 2017-02-20T01:24:23.366Z:
% 
% > This vision is too simplistic to render the intrinsic planning problem inherent to dialogue as well as its grounded nature
% What does this mean?
% 
% ^ <florian.strub@gmail.com> 2017-02-20T02:32:14.063Z:
%
% I think that Olivier wanted to say that supervised approaches would fail to learn long-term dependency strategies. 
% I would suggest : 
% This vision can be too simplistic since dialogue strategies require planning, i.e. sequential decision making, and errors may easily compound.
%
% ^.
\end{abstract}

\section{Introduction}
Ever since the formulation of the Turing Test, building systems that can meaningfully converse with humans has been a long-standing goal of Artificial Intelligence (AI). 
% First impressive examples of conversational agents included rule-based methods such as ELIZA~\cite{Weizenbaum:1966:ECP:365153.365168}. 
Practical dialogue systems have to implement a management strategy that defines the system's behavior, for instance to decide when to provide information or to ask for clarification from the user. Although traditional approaches use linguistically motivated rules~\cite{Weizenbaum:1966:ECP:365153.365168}, recent methods are data-driven and make use of Reinforcement Learning (RL)~\cite{lemon2007machine}. Significant progress in Natural Language Processing via Deep Neural Nets~\cite{bengio2003neural} made neural encoder-decoder architectures a promising way to train conversational agents ~\cite{vinyals2015neural,sordoni2015neural,serban2016generative}. The main advantage of such end-to-end dialogue systems is that they make no assumption about the application domain and are simply trained in a supervised fashion from large text corpora~\cite{lowe2015ubuntu}. 
\begin{figure}[t]
\centering
\includegraphics[width=1\linewidth]{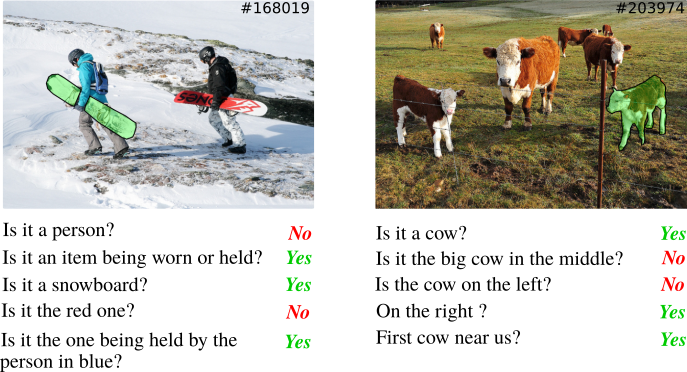}
\caption{Two example games of the \GW dataset. The correct object is highlighted by a green  mask. }
\label{fig:examples}
\vskip -1em
\end{figure}

However, there are many drawbacks to this approach. First, encoder-decoder models cast the dialogue problem into one of supervised learning, predicting the distribution over possible next utterances given the discourse so far. As with machine translation, this may result in inconsistent dialogues and errors that can accumulate over time. This is especially true because the action space of dialogue systems is vast, and existing datasets cover only a small subset of all trajectories, making it difficult to generalize to unseen scenarios~\cite{mooney2006learning}. Second, the supervised learning framework does not account for the intrinsic planning problem that underlies dialogue, \textit{i.e.} the sequential decision making process, which makes dialogue consistent over time. This is especially true when engaging in a task-oriented dialogue. As a consequence, reinforcement learning has been applied to dialogue systems since the late 90s~\cite{levin1997learning,Singh1999} and dialogue optimization has been generally more studied than dialogue generation. Third, it doesn't naturally integrate external contexts (larger than the history of the dialogue) that is most often used by dialogue participants to interact. This context can be their physical environment, a common task they try to achieve, a map on which they try to find their way, a database they want to access \textit{etc}. It is part of the so called \textit{Common Ground}, well studied in the discourse literature~\cite{COGS:COGS73}. Over the last decades, the field of cognitive psychology has also brought empirical evidence that human representations are grounded in perception and motor systems~\cite{barsalou2008grounded}. These theories imply that a dialogue system should be grounded in a multi-modal environment in order to obtain human-level language understanding~\cite{DBLP:journals/corr/KielaBVC16}. Finally, evaluating dialogues is difficult as there is not an automatic evaluation metric that correlates well with human evaluations~\cite{liu2016not}.
% * <aaron.courville@gmail.com> 2017-02-20T01:45:51.373Z:
% 
% > However, there are many drawbacks to this approach
% Consider transforming this paragraph into a numbered bullet list.
% 
% ^.
% * <aaron.courville@gmail.com> 2017-02-20T01:38:45.195Z:
% 
% > This is the reason why RL was part of the game since the late 90's~
% Please try to avoid colloquial language such as this. I can't fix it because I really don't understand what is really intended here.
% 
% ^.
% * <aaron.courville@gmail.com> 2017-02-20T01:36:53.398Z:
% 
% > As with machine translation
% Do we have a citation for this?
% 
% ^.

On the other hand, RL approaches could handle the planning and the non-differentiable metric problems but require online learning (although batch learning is possible but difficult with low amounts of data~\cite{pietquin2011sample}). For that reason, user simulation has been proposed to explore dialogue strategies in a RL setting~\cite{eckert1997user,schatzmann2006survey,pietquin2013}. It also requires the definition of an evaluation metric which is most often related to task completion and user satisfaction~\cite{walker1997paradise}. In addition, successful applications of the RL framework to dialogue often rely on a predefined structure of the task, such as slot-filling tasks~\cite{williams2007partially} where the task can be casted as filling in a form. 
%Typical examples of such systems include reserving a restaurant table\cite{wen2016network} or recommending a movie\cite{}. 

%The dialogue manager selects slots that will be used to query an external database in order to retrieve instances satisfying the user constraints. The system's utterance is then generated from dialogue state and the query results, often by selecting predefined templates\cite{}. The key advantage of slot-filling methods is that they significantly reduce the dialogue state space, and, therefore, require less data to train an effective system. However, slots are domain-specific and do not easily generalize to novel domains. Moreover, interacting with an external database is non-differentiable, and effectively incorporating it into an end-to-end framework is ongoing research\cite{}.   

In this paper, we present a global architecture for end-to-end RL optimization of a task-oriented dialogue system and its application to a multimodal task, grounding the dialogue into a visual context. To do so, we start from a corpus of 150k human-human dialogues collected via the recently introduced \GW game~\cite{hvries2016}. The goal of the game is to locate an unknown object in a natural picture by asking a series of questions. This task is hard since it requires scene understanding and, more importantly, a dialogue strategy that leads to identify the object rapidly. Out of these data, we first build a supervised agent and a neural training environment. It serves to train a DeepRL agent online which is able to solve the task. We then quantitatively and qualitatively compare the performance of our system to a supervised approach on the same task from a human baseline perspective. 
\begin{figure}[t]
\includegraphics[width=\linewidth]{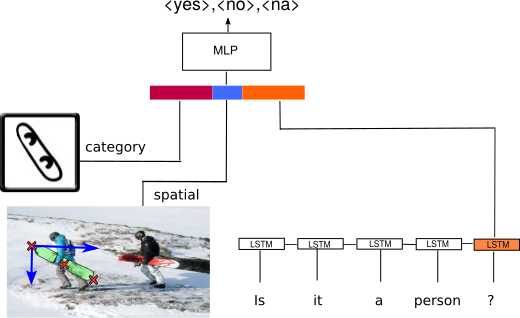}
\caption{Oracle model.}
\label{fig:oracle_model}
\end{figure}
%Previous work\cite{hvries2016} has focused on learning to generate questions in a purely supervised way. In this paper, we introduce a deep reinforcement learning framework to train an agent to generate a series of questions that successfully locates the object. 
In short, our contributions are:
\begin{itemize}
\item to propose the first multimodal goal-directed dialogue system optimized via Deep RL; 
\item to achieve 10\% improvement on task completion over a supervised learning baseline.
%\item to qualitatively analyze how dialogue strategy trained in a RL fashion may outperform supervised procedures
\end{itemize}

%The dialogue management system must interact with either the user or a simulated user \cite{eckert1998automatic}. Therefore, the dialogue manager may experience unknown dialogue space and it can potentially improve its strategy. 
%n the simulation-based setup, the dialogue manager is first pre-trained on the dialogue corpus in a supervised fashion. Then, it interacts with the simulated user to improve his policy by using Reinforcement Learning approach.  

\section{\GW game}

\begin{figure}[t]
\includegraphics[width=\linewidth]{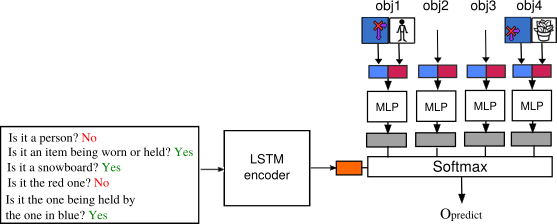}
\caption{Guesser model.}
\label{fig:guesser_model}
\end{figure}

We briefly explain here the \GW game that will serve as a task for our dialogue system, but refer to~\cite{hvries2016} for more details regarding the task and the exact content of the dataset. It is composed of more than 150k human-human dialogues in natural language collected through Mechanical Turk. 

\subsection{Rules}
\GW is a cooperative two-player game in which both players see the picture of a rich visual scene with several objects. One player -- the oracle -- is randomly assigned an object (which could be a person) in the scene.  This object is not known by the other player -- the questioner -- whose goal is to locate the hidden object. To do so, the questioner can ask a series of yes-no questions which are answered by the oracle as shown in  Fig~\ref{fig:examples}. Note that the questioner is not aware of the list of objects and can only see the whole picture. Once the questioner has gathered enough evidence to locate the object, he may choose to guess the object. The list of objects is revealed, and if the questioner picks the right object, the game is considered successful.

\subsection{Notation}\label{sec:notation}

Before we proceed, we establish the \GW notations that are used throughout the rest of this paper. A game is defined by a tuple $(\Image, D, O, o^*)$ where $\Image \in \mathbb{R}^{H \times W}$ is a picture of height $H$ and width $W$, $D$ a dialogue with $J$ question-answer pairs $D=(\bm{q}_j,a_j)_{j=1}^J$, $O$ a list of $K$ objects $O=(o_k)_{k=1}^K$ and $o^*$ the target object. Moreover, each question $\bm{q}_j = (w^j_i)_{i=1}^{I_j}$ is a sequence of length $I_j$ with each token $w^j_i$ taken from a predefined vocabulary $V$. The vocabulary $V$ is composed of a predefined list of words, a question tag $\stoptoken$ that ends a question and a stop token $\stopdialoguetoken $ that ends a dialogue. An answer is restricted to be either yes, no or not applicable \textit{i.e}. $a_j \in \{\yestoken, \notoken, \natoken\}$.  For each object $k$, an object category $c_k \in \{1, \dots, C\}$ and a pixel-wise segmentation mask $S_k \in \{0, 1\}^{H \times W}$ are available.
% * <aaron.courville@gmail.com> 2017-02-20T02:35:47.372Z:
% 
% >  $\bm{q}_j = (w^j_i)_{i=1}^{I_j}$
% I find this notation quite confusing. Please carefully walk through each element of the notation.
% 
% ^.
Finally, to access subsets of a list, we use the following notations. If $l=(l_i^j)_{i=1}^{I,j}$ is a double-subscript list, then $l^{j}_{\ft{1}{i}}=(l_p^j)_{p=1}^{i,j}$ are the $i$ first elements of the $j^{th}$ list if $1\leq i\leq I_j$, otherwise $l^{j}_{\ft{1}{p}}=\emptyset$. 
Thus, for instance, $w^j_{\ft{1}{i}}$ refers to the first $i$ tokens of the $j^{th}$ question and $(\bm{q},a)_{1:j}$ refers to the $j$ first question-answer pairs of a dialogue.
%Old
%Finally, to access subsets of a list, we use the following notations. If $l=(l_i^j)_{i=1,j=1}^{N,M}$ is a double-subscript list, then $l^{\ft{1}{q}}_{\ft{1}{p}}=(l_i^j)_{i=1,j=1}^{p,q}$ if $1\leq p\leq N$ and $1\leq q\leq M$, otherwise $l^{\ft{1}{q}}_{\ft{1}{p}}=\emptyset$. If $l=(l_i)_{i=1}^N$ is a list of length N, then $l_{\ft{1}{j}}=(l_i)_{i=1}^j$ only if $1\leq j\leq N$, otherwise $l_{\ft{1}{j}}=\emptyset$. Thus, for instance, $w^{\ft{1}{q}}_{\ft{1}{p}}$ refers to the first $p$ tokens of the first $q$ questions.

\begin{figure}[t]
\includegraphics[width=\linewidth]{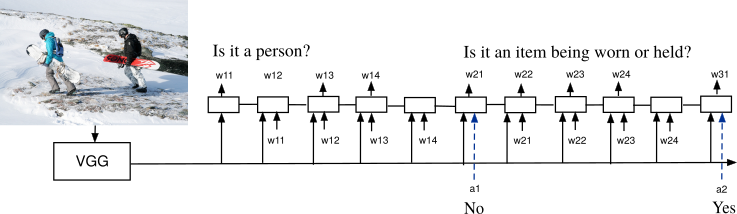}
\vskip -1em 
\caption{Question generation model.}
\label{fig:qgen_model}
\vskip -0.5em
\end{figure}

\section{Training environment}\label{sec:models}
From the \GW dataset, we build a training environment that allows RL optimization of the questioner task by creating models for the oracle and guesser tasks. We also describe the supervised learning baseline to which we will compare. This mainly reproduces baselines introduced in~\cite{hvries2016}.  
%Because our framework relies on these models to build the simulation environment, we briefly review the tasks and its neural architectures that achieved the best performance. 

\paragraph{Question generation baseline}
We split the questioner's job into two different tasks: one for asking the questions and another one for guessing the object. The question generation task requires to produce a new question $\bm{q}_{j+1}$, given an image $\Image$ and a history of $j$ questions and answers $(\bm{q},a)_{1:j}$. We model the question generator (QGen) with a recurrent neural network (RNN), which produces a sequence of RNN state vectors $\bm{s}^j_{1:i}$ for a given input sequence $\bm{w}^j_{1:i}$ by applying the transition function $f$:
$
\bm{s}^j_{i+1} = f(\bm{s}^j_i, w^j_i).
$
We use the popular long-short term memory (LSTM) cell \cite{hochreiter1997long} as our transition function. In order to construct a probabilistic sequence model, one can add a softmax function $g$ that computes a distribution over tokens $w^j_i$ from vocabulary $V$. In the case of \GW, this output distribution is conditioned on all previous questions and answers tokens as well as the image $\Image$:
\begin{equation}
 p(w^j_{i} |w^j_{\ft{1}{i-1}}, (\bm{q},a)_{\ft{1}{j-1}} , \Image).
\end{equation}
We condition the model on the image by obtaining its VGG16 FC8 features and concatenating it to the input embedding at each step, as illustrated in Fig.~\ref{fig:qgen_model}. We train the model by minimizing the conditional negative log-likelihood:
\begin{multline*}
- \log p(\bm{q}_{\ft{1}{J}} | a_{\ft{1}{J}}, \Image) = - \log \prod_{j=1}^J p(\bm{q}_{j} | (\bm{q},a)_{\ft{1}{j-1}}, \Image),
\\ = - \sum_{j=1}^{J} \sum_{i=1}^{I_j} \log p(w^j_i | w^j_{\ft{1}{i-1}}, (\bm{q},a)_{\ft{1}{j-1}}, \Image).
\end{multline*}
At test time, we can generate a sample $p(\bm{q}_{j} | (\bm{q},a)_{\ft{1}{j-1}}, \Image)$ from the model as follows. Starting from the state $\bm{s}^j_{1}$, we sample a new token $w^j_i$ from the output distribution $g$ and feed the embedded token $e(w^j_i)$ back as input to the RNN. We repeat this loop till we encounter an end-of-sequence token. To approximately find the most likely question, $\text{max}_{\bm{q}_j}\ p(\bm{q}_{j} | (\bm{q},a)_{\ft{1}{j-1}}, \Image)$, we use the commonly used beam-search procedure. This heuristics aims to find the most likely sequence of words by exploring a subset of all questions and keeping the $K$-most promising candidate sequences at each time step.

\paragraph{Oracle}
The oracle task requires to produce a yes-no answer for any object within a picture given a natural language question. We outline here the neural network architecture that achieved the best performance and refer to \cite{hvries2016} for a thorough investigation of the impact of other object and image information. First, we embed the spatial information of the crop by extracting an 8-dimensional vector of the location of the bounding box $[ x_{min}, y_{min}, x_{max}, y_{max},x_{center}, y_{center}, w_{box}, h_{box}]$
where $w_{box}$ and $h_{box}$ denote the width and height of the bounding box
, respectively. We normalize the image height and width such that coordinates range from $-1$ to $1$, and place the origin at the center of the image. 
Second, we convert the object category $c^*$ into a dense category embedding using a learned look-up table. Finally, we use a LSTM to encode the current question $q$. We then concatenate all three embeddings into a single vector and feed it as input to a single hidden layer MLP that outputs the final answer distribution $p(a|q, c^*, x^*_{spatial})$ using a softmax layer, illustrated in Fig.~\ref{fig:oracle_model}. 
 %Finally, we use ADAM to minimize the cross-entropy error for $20$ epochs, and use early stopping on the validation error. 

\paragraph{Guesser}
The guesser model takes an image $\Image$ and a sequence of questions and answers $(\bm{q},a)_{\ft{1}{N}}$, and predicts the correct object $o^*$ from the set of all objects. This model considers a dialogue as one flat sequence of question-answer tokens and use the last hidden state of the LSTM encoder as our dialogue representation. We perform a dot-product between this representation and the embedding for all the objects in the image, followed by a softmax to obtain a prediction distribution over the objects. The object embeddings are obtained from the categorical and spatial features. More precisely, we concatenate the 8-dimensional spatial representation and the object category look-up and pass it through an MLP layer to get an embedding for the object. Note that the MLP parameters are shared to handle the variable number of objects in the image. See Fig~\ref{fig:guesser_model} for an overview of the guesser. 

\subsection{Generation of full games}
With the question generation, oracle and guesser model we have all components to simulate a full game. Given an initial image $\Image$, we generate a question $\bm{q}_1$ by sampling tokens from the question generation model until we reach the question-mark token. Alternatively, we can replace the sampling procedure by a beam-search to approximately find the most likely question according to the generator. The oracle then takes the question $\bm{q}_1$, the object category $c^*$ and $x_{spatial}^*$ as inputs, and outputs the answer $a_1$. We append $(\bm{q}_1, a_1)$ to the dialogue and repeat generating question-answer pairs until the generator emits a stop-dialogue token or the maximum number of question-answers is reached. Finally, the guesser model takes the generated dialogue $D$ and the list of objects $O$ and predicts the correct object.

% TODO reduce to a few sentences
% \subsection{Short-comings of imitation learning}
% First, all the negative output are equally penalized which can reduce the granularity of the training. More importantly, this approach can only be applied over a predefined dataset. It thus assumes that the dataset distribution covers the full (or at least sufficient) space of dialogues. While this assumption may be fairly reasonable in some translation corpus~\cite{bojar2014findings}, it completely breaks down in the dialogue setting~\cite{schatzmann2006survey,lemon2007machine}. These authors then advocate that a dialogue model must explore the space of dialogues by having positive and negative feedback. This reinforcement learning setting allows the dialogue system to learn new probability distributions that may not exist in the original dataset.

\section{\GW from RL perspective}
One of the drawbacks of training the QGen in a supervised learning setup is that its sequence of questions is not explicitly optimized to find the correct object. Such training objectives miss the planning aspect underlying (goal-oriented) dialogues. In this paper, we propose to cast the question generation task as a RL task. More specifically, we use the training environment described before and consider the oracle and the guesser as part of the RL agent environment. In the following, we first formalize the \GW task as a Markov Decision Process (MDP) so as to apply a policy gradient algorithm to the QGen problem. 

\subsection{\GW as a Markov Decision Process}
We define the state $\bm{x_t}$ as the status of the game at step $t$. Specifically, we define $\bm{x_t} = ((w^j_1, \ldots, w^j_{i}), (\bm{q},a)_{\ft{1}{j-1}}, \Image)$ where $t=\sum_{j=1}^{j-1}I_j + i$ corresponds to the number of tokens generated since the beginning of the dialogue. An action $u_t$ corresponds to select a new word $w^j_{i+1}$ in the vocabulary $V$. The transition to the next state depends on the selected action:
\begin{itemize}[noitemsep,leftmargin=*]
\item If $w^j_{i+1} = \stopdialoguetoken$, the full dialogue is terminated.
\item If $w^j_{i+1} = \stoptoken$, the ongoing question is terminated and an answer $a_j$ is sampled from the oracle. The next state is $\bm{x}_{t+1} = ((\bm{q},a)_{\ft{1}{j}}, \Image)$ where $\bm{q}_j = (w^j_{1}, \ldots, w^j_{i}, \stoptoken)$. 
\item Otherwise  the new word is appended to the ongoing question and $\bm{x}_{t+1} = ((w^j_{1}, \ldots, w^j_{i}, w^j_{i+1}), (\bm{q},a)_{\ft{1}{j-1}}, \Image)$. 
\end{itemize}
Questions  are automatically terminated after $I_{max}$ words. Similarly, dialogues are terminated after $J_{max}$ questions. Furthermore, a reward $r(\bm{x},u)$ is defined for every state-action pair. 
A trajectory $\tau = (\bm{x}_t,u_t,\bm{x}_{t+1}, r(\bm{x}_t,u_t))_{\ft{1}{T}}$ is a finite sequence of tuples of length $T$ which contains a state, an action, the next state and the reward where $T \leq J_{max}*I_{max}$. Thus, the game falls into the episodic RL scenario as the dialogue terminates after a finite sequence of question-answer pairs. Finally, the QGen output can be viewed as a stochastic policy  $\pi_{\bm{\theta}}(u|\bm{x})$  parametrized by $\bm{\theta}$ which associates a probability distribution over the actions (i.e. words) for each state (\textit{i.e.} intermediate dialogue and picture). 
 
\subsection{Training the QGen with Policy Gradient}
While several approaches exist in the RL literature, we opt for policy gradient methods because they are known to scale well to large action spaces. This is especially important in our case because the vocabulary size is nearly 5k words. 
The goal of policy optimization is to find a policy $\pi_{\bm{\theta}}(u|\bm{x})$ that maximizes the expected return, also known as the mean value:
\begin{equation}
J(\bm{\theta}) = E_{\pi_{\bm{\theta}}}\big[\sum^T_{t=1}\gamma^{t-1}r(\bm{x}_t,u_t) \big],
\end{equation} 
where $\gamma \in [0,1]$ is the discount factor, $T$ the length of the trajectory and the starting state $x_1$ is drawn from a distribution $p_1$. Note that $\gamma=1$ is allowed as we are in the episodic scenario~\cite{sutton1999policy}. To improve the policy, its parameters can be updated in the direction of the gradient of the mean value:
\begin{equation}
\bm{\theta}_{h+1} = \bm{\theta}_{h} + \alpha_h \nabla_\theta J|_{\theta=\theta_h},
\end{equation} 
where $h$ denotes the training time-step and $\alpha_h$ is a learning rate such that $\sum_{h=1}^{\infty} \alpha_h =  \infty$ and $\sum_{h=1}^{\infty} \alpha_h^2 < \infty$. 

Thanks to the gradient policy theorem~\cite{sutton1999policy}, the gradient of the mean value can be estimated from a batch of trajectories $\mathcal{T}_h$ sampled from the current policy $\pi_{\bm{\theta}_h}$ by:    
\begin{equation}
\label{eq:full_policy_gradient}
\nabla J(\bm{\theta}_h) = \bigg \langle
\sum_{t=1}^{T} \sum_{u_t\in V} \nabla_{\bm{\theta}_h}\text{log }{\pi_{\bm{\theta}_h}}(u_t|\bm{x}_t)(Q^{\pi_{\bm{\theta}_h}}(\bm{x}_t, u_t) - b) \bigg \rangle_{\mathcal{T}_h},
\end{equation}
where $Q^{\pi_{\bm{\theta}_h}}(\bm{x}, u)$ is the state-action value function that estimates the cumulative expected reward for a given state-action couple and $b$ some arbitrarily baseline function which can help reducing the variance of the estimation of the gradient. More precisely 
\begin{equation}
Q^{\pi_{\bm{\theta}_h}}(\bm{x}_t, u_t)= E_{\pi_{\bm{\theta}}}\big[\sum^T_{t'=t}\gamma^{t'-t}r(\bm{x}_{t'},u_{t'})\big].
\end{equation}

Notice that the estimate in Eq~\eqref{eq:full_policy_gradient} only holds if the probability distribution of the initial state $\bm{x_1}$ is uniformly distributed. The state-action value-function $Q^{\pi_{\bm{\theta}_h}}(\bm{x}, u)$ can then be estimated by either learning a function approximator (Actor-critic methods) or by Monte-Carlo rollouts (REINFORCE~\cite{williams1992simple}). In REINFORCE, the inner sum of actions is estimated by using the actions from the trajectory. Therefore, Eq~\eqref{eq:full_policy_gradient} can be simplified to:    
\begin{equation}
\label{eq:policy_gradient}
\nabla J(\bm{\theta}_h) = \bigg \langle
\sum_{t=1}^{T} \nabla_{\bm{\theta}_h}\text{log }{\pi_{\bm{\theta}_h}}(u_t|\bm{x}_t)(Q^{\pi_{\bm{\theta}_h}}(\bm{x}_t, u_t) - b) \bigg \rangle_{\mathcal{T}_h}.
\end{equation}
Finally, by using the \GW game notation for Eq~\eqref{eq:policy_gradient}, the policy gradient for the QGen can be written as:
\begin{multline}
\label{eq:gw_policy_gradient}
\nabla J(\bm{\theta}_h) = \bigg \langle
\sum_{j=1}^J \sum_{i=1}^{I_j} \nabla_{\bm{\theta}_h}\text{log }{\pi_{\bm{\theta}_h}}(w^j_i|w^{j}_{\ft{1}{i-1}},(\bm{q},a)_{\ft{1}{j-1}},  \Image)  \\ (Q^{\pi_{\bm{\theta}_h}}((w^{j}_{\ft{1}{i-1}},(\bm{q},a)_{\ft{1}{j-1}}, \Image), w^j_i) - b) \bigg \rangle_{\mathcal{T}_h}.
\end{multline}
%This policy gradient can be put in parallel to similar approaches in the translation community~\cite{liu2016optimization,bahdanau2016actor,ranzato2015sequence}. %TODO F find a better conclusion!

\subsection{Reward Function}
One tedious aspect of RL is to define a correct and valuable reward function. As the optimal policy is the result of the reward function, one must be careful to design a reward that would not change the expected final optimal policy~\cite{ng1999policy}. Therefore, we put a minimal amount of prior knowledge into the reward function and construct a zero-one reward depending on the guesser's prediction:
\begin{equation}
  r(\bm{x}_t, u_t)=\begin{cases}
    1 & \text{If argmax}_o[Guesser(\bm{x}_t)]=o* \text{ and } t=T \\
    0 & \text{Otherwise}\\
    \end{cases}.
\end{equation}
So, we give a reward of one if the correct object is found from the generated questions, and zero otherwise. 

Note that the reward function requires the target object $o*$ while it is not included in the state $\bm{x} = ((\bm{q},a)_{\ft{1}{J}}, \Image)$. This breaks the MDP assumption that the reward should be a function of the current state and action. However, policy gradient methods, such as REINFORCE, are still applicable if the MDP is partially observable~\cite{williams1992simple}.

\subsection{Full training procedure}
For the QGen, oracle and guesser, we use the model architectures outlined in section \ref{sec:models}. We first independently train the three models with a cross-entropy loss. We then keep the oracle and guesser models fixed, while we train the QGen in the described RL framework. It is important to pretrain the QGen to kick-start training from a reasonable policy. The size of the action space is simply too big to start from a random policy.

In order to reduce the variance of the policy gradient, we implement the baseline $b_{\bm{\phi}}(\bm{x}_t)$ as a function of the current state, parameterized by $\bm{\phi}$. Specifically, we use a one layer MLP which takes the LSTM hidden state of the QGen and predicts the expected reward. We train the baseline function by minimizing the Mean Squared Error (MSE) between the predicted reward and the discounted reward of the trajectory at the current time step:  
\begin{equation}
\label{eq:baseline_mse}
L(\bm{\phi_h}) = \Big \langle \big[ b_{\phi_h}(\bm{x}_t) - \sum_{t'=t}^T\gamma^{t'} r_{t'} \big]^2  \Big \rangle_{\mathcal{T}_h}
\end{equation}

We summarize our training procedure in Algorithm \ref{reinforce}. 

\begin{algorithm}[t]
 \begin{algorithmic}[1] 
 \Require Pretrained QGen,Oracle and Guesser
 \Require Batch size $K$
%\Procedure{QGen Training loop}{}
	\For{Each update}
	\State \# Generate trajectories $\mathcal{T}_h$
    \For{$k = 1 \text{ to } K$}
    \State Pick Image $\Image_k$ and the target object $o_k^* \in O_ k$
    \State \# Generate question-answer pairs $(\bm{q},a)^k_{1:j}$ 
	\For{$j = 1 \text{ to } J_{max}$}
        \State $q^k_j = QGen(\bm{q},a)^k_{1:j-1},\Image_k)$
        \State $a^k_j = Oracle(\bm{q}^k_j, o_k^*, \Image_k)$
        \If{$\stopdialoguetoken \in \bm{q}^k_j$ }
           \State delete $(q,a)^k_j$ and break;
        \EndIf
	\EndFor 
    \State $p(o_k | \cdot) = Guesser((q,a)^k_{1:j}, \Image_k, O_k)$
    \State $r(\bm{x}_t, u_t)=\begin{cases}
    1 & \text{If argmax}_{o_k} p(o_k | \cdot)=o_k^*\\
    0 & \text{Otherwise}\\
    \end{cases}$
    \EndFor
    \State Define $\mathcal{T}_h = ((q,a)^k_{1:j_k}, \Image_k, r_k)_{1:K}$
    \State Evaluate $\nabla J(\bm{\theta}_h)$ with Eq.~\eqref{eq:gw_policy_gradient} with $\mathcal{T}_h$
        \State SGD update of QGen parameters $\bm{\theta}$ using $\nabla J(\bm{\theta}_h)$
     \State Evaluate $\nabla L(\bm{\phi}_h)$ with Eq.~\eqref{eq:baseline_mse} with $\mathcal{T}_h$
    \State SGD update of baseline parameters using  $\nabla L(\bm{\phi}_h)$ %with
    
%\EndProcedure
\EndFor
\end{algorithmic}
\caption{Training of QGen with REINFORCE}
\label{reinforce}
\end{algorithm}

%In the \GW game, the conversational agent is defined as the questioner while the user simulation corresponds to the oracle. Given a picture, the questioner first asks a question, that will be answered by the oracle. The dialogue consists in repeating this interaction until a given criterion is reached (maximum number of question, confidence threshold to guess the object etc.). Therefore, the oracle can be solely trained on the \GW dataset to correctly answer the other player's questions. 

% \begin{figure}[t]
% \centering
% \begin{subfigure}{0.48\linewidth}
% \includegraphics[width=\linewidth]{sampling_model/model_sampling_QuestionVsDialogue}
% \caption{}
% \label{fig:number_question}
% \end{subfigure}
% \begin{subfigure}{0.48\linewidth}
% \includegraphics[width=\linewidth]{sampling_model/model_sampling_SuccessDialogueLength}
% \caption{}
% \label{fig:length_success}
% \end{subfigure}
% \centering
% \caption{(a) Number of question per dialogue (b) Success ratio per length of dialogues.}
% \end{figure}

\begin{figure}[t]
\centering
\includegraphics[width=0.7\linewidth]{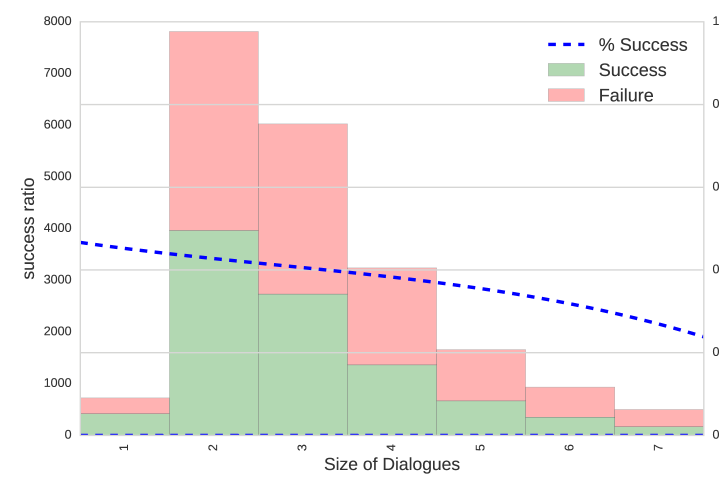}
\vskip -1em
\caption{Task completion ratio of REINFORCE trained QGEN for given dialogue length.}
\label{fig:length_success}
\vskip -1em
\end{figure}

\section{Related work}

Outside of the dialogue literature, RL methods have been applied to encoder-decoder architectures in machine translation~\cite{ranzato2015sequence,bahdanau2016actor} and image captioning~\cite{liu2016optimization}. In those scenarios, the BLEU score is used as a reward signal to fine-tune a network trained with a cross-entropy loss. However, the BLEU score is a surrogate for human evaluation of naturalness, so directly optimizing this measure does not guarantee improvement in the translation/captioning quality. In contrast, our reward function encodes task completion, and optimizing this metric is exactly what we aim for. Finally, the BLEU score can only be used in a batch setting because it requires the ground-truth labels from the dataset. In \GW, the computed reward is independent from the \emph{human} generated dialogue. 
%RL is mainly used to circumvent the  non-differentiabl aspect of Blue score as it cannot be directly optimized by Gradient descent.
% Noticeably, the fully observable MDP assumptions is also broken while computing the BLUE score as it requires some hidden ground samples state to be computed (which of course, are not in the current state).  
%Differently, adversarial networks have been used to compute a data-independent reward signal~\cite{kannan2017adversarial} to generate natural language utterance in a RL fashion, but it is still ongoing research. 

Although visually-grounded language models have been studied for a long time~\cite{roy2002learning}, important breakthroughs in both visual and natural language understanding has led to a renewed interest in the field~\cite{lecun2015deep}. Especially image captioning~\cite{lin2014microsoft} and visual question answering~\cite{antol2015vqa} has received much attention over the last few years, and encoder-decoder models~\cite{liu2016optimization,lu2016hierarchical} have shown promising results for these tasks. Only very recently the language grounding tasks have been extended to a dialogue setting with the Visual Dialog~\cite{das2016visual} and \GW~\cite{hvries2016} datasets. While Visual Dialog considers the chit-chat setting, the \GW game is goal-oriented which allows us to cast it in into an RL framework. %Differently, \cite{geman2015visual}
%malinowski2015ask

% which may greatly reduce the scope of potential virtual embodiment~\cite{mooney2006learning,DBLP:journals/corr/KielaBVC16}. As these visual dialogue datasets have just been released, the field is still under active ongoing research.

\section{Experiments}

\begin{table*}
\scriptsize 
\begin{tabular}{m{2.6cm}| m{2.7cm} | m{2.3cm} | m{2.6cm}| m{3.0cm} | m{2.4cm}}
\makecell{\textbf{Image}} & \makecell{\textbf{{\color{cyan}Beam Search}}} & \makecell{\textbf{{\color{magenta}REINFORCE}}} & \makecell{\textbf{Image}} & \makecell{\textbf{{\color{cyan}Beam Search}}} & \makecell{\textbf{{\color{magenta}REINFORCE}}}\\
\hline
\multirow{7}{*}{\makecell{\includegraphics[width=2.8cm]{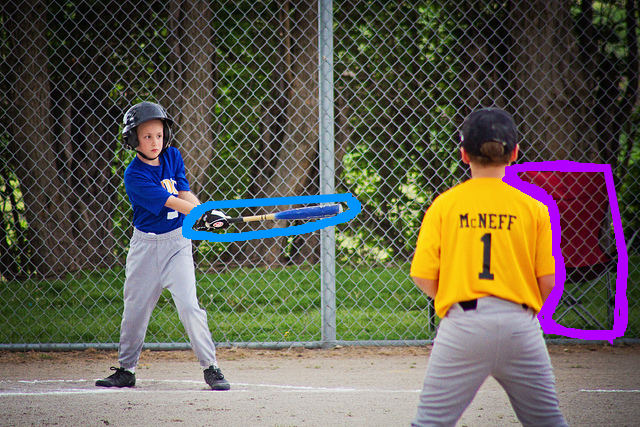}}}
& Is it a person ? no & Is it a person ? no & \multirow{7}{*}{\makecell{\includegraphics[width=2.7cm]{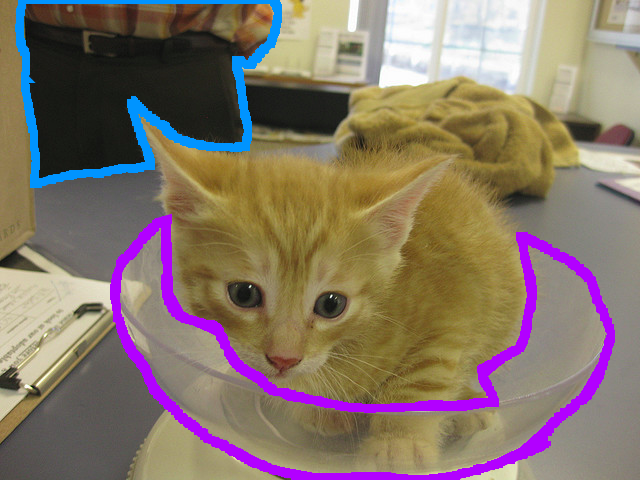}}} & Is it a cat ? no & Is it a cat ? no\\
& Is it a ball ? no & Is a glove ? no && Is it a book ? no & Is it on the table ? yes \\ 
& Is it a ball ? no & Is an umbrella ? no && Is it a book ? no & Is it the book ? no\\ 
& Is it a ball ? no & Is in the middle ? no && Is it a book ? no & Is it fully visible? yes \\
& Is it a ball ? no & On a person? no && Is it a book ? no &\\
& & is it on on far right? yes & & &\\ 
& \makecell{{\color{red}Failure} (blue bat)} & \makecell{{\color{green} Success} (red chair)} && \makecell{{\color{red}Failure} (person)} & \makecell{{\color{green} Success} (bowl)}\\
\hline 
\multirow{7}{*}{\;  \includegraphics[width=2.5cm]{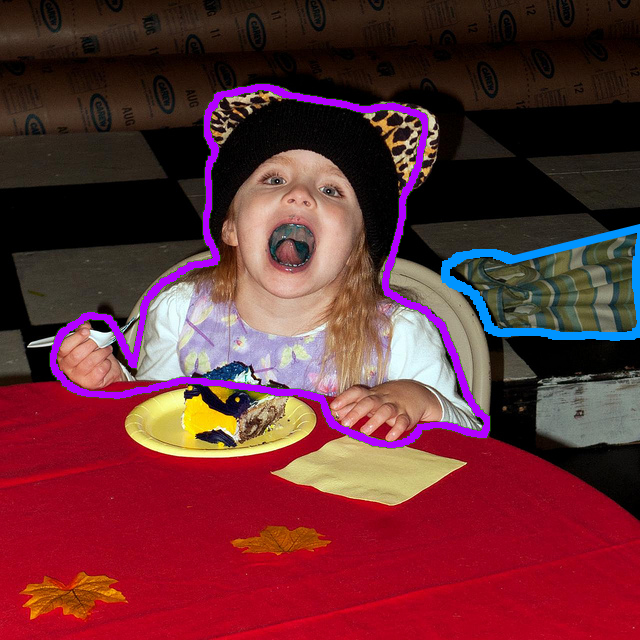}} 
& Is it a person ? yes & Is it a person ? yes &
\multirow{7}{*}{\makecell{\; \\\includegraphics[width=2.8cm]{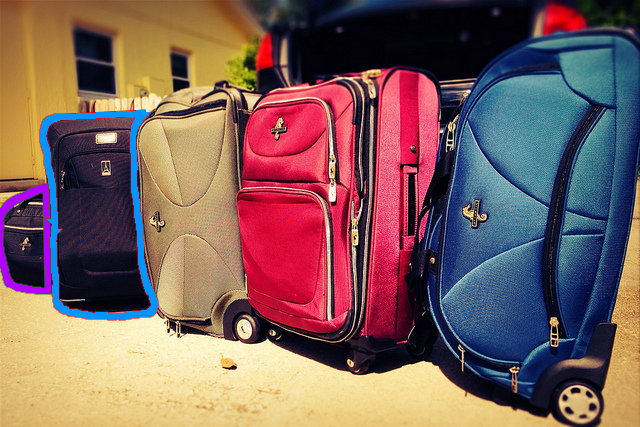} }} & Is it a bag ? yes & Is it a suitcase? yes \\
& Is it the one in front ? yes & Is it girl in white ? yes && Is it red ? no & Is it in the left side ? yes\\
& Is it the one on the left ? no &&& Is it the one in the middle ? no & \\
& Is it the one in the middle with the red umbrella ? yes &&& Is it the one on the far right ? no & \\
& Is it the one to the right of the girl in ? no & & & Is it the one with the blue bag ? yes& \\
& \makecell{{\color{red}Failure} (umbrella)} & \makecell{{\color{green} Success} (girl)} && \makecell{{\color{green} Success} (most left bag)} & \makecell{{\color{red}Failure} (left bag)}\\
& & & &\\ 
\hline
\end{tabular}
\caption{Samples extracted from the test set. The blue (resp. purple) box corresponds to the object picked by the guesser for the beam-search (resp. REINFORCE) dialogue. The small verbose description is added to refer to the object picked by the guesser.}
\vskip -1em
\label{table:samples} 
\end{table*}

As already said, we used the \GW dataset\footnote{Available at \url{https://guesswhat.ai/download}} that includes 155,281 dialogues containing 821,955 question/answer pairs composed of 4900 words on 66,537 unique images and 134,074 unique objects. The experiments source code is available at \url{https://guesswhat.ai}.

%\paragraph{Oracle}   

\subsection{Training details}
% The oracle and guesser are both based on a LSTM with 512 hidden neurons, a look-up table of dimension 512 for both object categories and word embeddings. Furthermore, the guesser uses a one layer MLP with 512 units to compute the object embedding. Finally, the QGen uses a LSTM with 1024 hidden units and a 512-dimensional word embedding. For each model, we use ADAM for optimization, we clip the gradient norm to 5 and train them for at most $30$ epochs. We use early stopping on the validation set, and report the test error. 
We pre-train the networks described in Section \ref{sec:models}. After training, the oracle network obtains $21.5\%$ error and the guesser network reports $36.2\%$ error on the test set. Throughout the rest of this section we refer to the pretrained QGen as our baseline model. 
We then initialize our environment with the pre-trained models and train the QGen with REINFORCE for 80 epochs with plain stochastic gradient descent (SGD) with a learning rate of 0.001 and a batch size of 64. For each epoch, we sample each training images once, and randomly choose one of its object as the target. We simultaneously optimize the baseline parameters $\bm{\phi}$ with SGD with a learning rate of $0.001$. Finally, we set the maximum number of questions to $8$ and the maximum number of words to $12$

\subsection{Results}
\paragraph{Accuracy} Since we are interested in human-level performance, we report the accuracies of the models as a percentage of human performance (84.4\%), estimated from the dataset. We report the scores in Table~\ref{table:results}, in which we compare sampling objects from the training set (New Objects) and test set (New Pictures) i.e. unseen pictures. We report the standard deviation over 5 runs in order to account for the sampling stochasticity. On the test set, the baseline obtains $45.0\%$ accuracy, while training with REINFORCE improves to $62.0\%$. This is also a significant improvement over the beam-search baseline, which achieves $53.0\%$ on the test-set. The beam-search procedure improves over sampling from the baseline, but interestingly lowers the score for REINFORCE.    

\begin{table}
\small 
\begin{tabular}{|l|r|r|r|}
  \hline
    \multicolumn{2}{|c|}{} & \makecell{New Objects} & \makecell{New Pictures} \\ 
  \hline 
  \multirow{3}{*}{Baseline} 
    & Sampling & 46.4\% $\pm$ 0.2 & 45.0\% $\pm$ 0.1\\
    & Greedy   & 48.2\% $\pm$ 0.1 & 46.9\% \\
    & BSearch  & 53.4\% $\pm$ 0.0 & 53.0\% \\ \hline
  \multirow{2}{*}{REINFORCE} 
    & Sampling & $\bm{63.2\%  \pm  0.3}$ & $\bm{62.0\%\pm 0.2}$ \\
    & Greedy   &      58.6\% $\pm$ 0.0   &      57.5\% \\
    & BSearch  &      54.3\% $\pm$ 0.1   &      53.2\% \\
    \hline
\end{tabular}
\caption{Accuracies of the models as a percentage of human performance of the QGen trained with the baseline and REINFORCE. New objects refers to uniformly sampling objects within the training set, while new pictures refer to the test set. }
\label{table:results}
\vskip -1em
\end{table}

\paragraph{Samples} 
We qualitatively compare the two methods by analyzing a few generated samples, as shown in Table \ref{table:samples}. We observe that the beam-search baseline trained in a supervised fashion keeps repeating the same questions, as can be seen in the two top examples in Tab.~\ref{table:samples}. We noticed this behavior especially on the test set \textit{i.e.} when confronted with unseen pictures, which may highlight some generalization issues. We also find that the beam-search baseline generates longer questions ($7.1$ tokens on average) compared to REINFORCE ($4.0$ tokens on average). This qualitative difference is clearly visible in the bottom-left example, which also highlights that the supervised baseline sometimes generates visually relevant but incoherent sequences of questions. For instance, asking "Is it the one to the right of the girl in?" is not a very logical follow-up of "Is it the one in the middle with the red umbrella?". In contrast, REINFORCE seem to implement a more grounded and relevant strategy: "Is it girl in white?" is a reasonable follow-up to "Is it a person?". In general, we observe that REINFORCE favor enumerating object categories ("is it a person?") or absolute spatial information ("Is it left?"). Note these are also the type of questions that the oracle is expected to answer correctly, hence, REINFORCE is able to tailor its strategy towards the strengths of the oracle.

\paragraph{Dialogue length} For the REINFORCE trained QGen, we investigate the impact of the dialogue length on the success ratio in Fig.~\ref{fig:length_success}. Interestingly, REINFORCE learns to stop on average after $4.1$ questions, although we did not encode a question penalty into the reward function. This policy may be enforced by the guesser since asking additional but noisy questions greatly lower the prediction accuracy of the guesser as shown in Tab.~\ref{table:samples}. Therefore, the QGen learns to stop asking questions when a dialogue contains enough information to retrieve the target object. However, we observe that the QGen sometimes stops too early, especially when the image contains too many objects of the same category. Interestingly, we also found that the beam-search fails to stop the dialogue. Beam-search uses a length-normalized log-likelihood to score candidate sequences to avoid a bias towards shorter questions. However, questions in \GW almost always start with "is it", which increases the average log likelihood of a question significantly. The score of a new question might thus (almost) always be higher than emitting a single $\stopdialoguetoken$ token. Our finding was further confirmed by the fact that a sampling procedure did stop the dialogue.

\paragraph{Vocabulary}
% * <aaron.courville@gmail.com> 2017-02-20T02:47:55.337Z:
% 
% > Vocabulary
% If possible, this paragraph should include more reflection of the significance of the results.
% 
% ^.
Sampling from the supervised baseline on the test set results in 2,893 unique words, while sampling from the REINFORCE trained model reduces its size to 1,194. However, beam search only uses 512 unique words which is consistent with the observed poor variety of questions. 

\vspace{-0.5em}
\section{Conclusion}
In this paper, we proposed to build a training environment from supervised deep learning baselines in order to train a DeepRL agent to solve a goal-oriented multi-modal dialogue task. We show the promise of this approach on the \GW dataset, and observe quantitatively and qualitatively an encouraging improvement over a supervised baseline model. While supervised learning models fail to generate a coherent dialogue strategy, our method learns when to stop after generating a sequence of relevant questions. 

%\item $R(s_t|g_t) = \gamma^{t} r_{N,T}$
%\item 
%$Q_\theta(s_t|g_t) = \frac{1}{K}\sum_{k=1}^{K} r_k$ with $r_k$ is computed by doing a rollout from $w_{t,1:T-1})$  

\paragraph{Acknowledgement} The authors would like to acknowledge the stimulating environment provided by the SequeL labs. We acknowledge the following agencies for research funding and computing support: CHISTERA IGLU and CPER Nord-Pas de Calais/FEDER DATA Advanced data science and technologies 2015-2020, NSERC, Calcul Qu\'{e}bec, Compute Canada, the Canada Research Chairs and CIFAR.

{\small
\bibliography{biblio}
\bibliographystyle{biblio_style}
}

\end{document}